\documentclass[conference]{IEEEtran}
\IEEEoverridecommandlockouts
\usepackage{cite}
\usepackage{amsmath,amssymb,amsfonts}
\usepackage{algorithmic}
\usepackage{graphicx}
\usepackage{textcomp}
\usepackage{xcolor}
\usepackage{hyperref}
\usepackage{tabularx}
\usepackage{multirow}

\newcolumntype{Y}{>{\centering\arraybackslash}X}

\def\BibTeX{{\rm B\kern-.05em{\sc i\kern-.025em b}\kern-.08em
    T\kern-.1667em\lower.7ex\hbox{E}\kern-.125emX}}
    
\begin{document}

\title{{\small{}9th International Conference on Research in Air Transportation (ICRAT 2020)}{\footnotesize{} }\\
    Model Generalization in Arrival Runway Occupancy Time Prediction by Feature Equivalences
}

% Example author bloc
% \author{
%     \IEEEauthorblockN{
%         Michael Shell\IEEEauthorrefmark{1}, 
%         Homer Simpson\IEEEauthorrefmark{2}, 
%         James Kirk\IEEEauthorrefmark{3}, 
%         Montgomery Scott\IEEEauthorrefmark{3},
%         and Eldon Tyrell\IEEEauthorrefmark{4}
%     }
%     \IEEEauthorblockA{
%         \IEEEauthorrefmark{1}
%         School of Electrical and Computer Engineering\\
%         Georgia Institute of Technology, Atlanta, Georgia 30332--0250\\
%         Email: mshell@ece.gatech.edu}
%     \IEEEauthorblockA{
%         \IEEEauthorrefmark{2}
%         Twentieth Century Fox, Springfield, USA\\
%         Email: homer@thesimpsons.com}
%     \IEEEauthorblockA{
%         \IEEEauthorrefmark{3}
%         Starfleet Academy, San Francisco, California 96678-2391\\
%         Telephone: (800) 555--1212, Fax: (888) 555--1212}
%     \IEEEauthorblockA{
%         \IEEEauthorrefmark{4}
%         Tyrell Inc.,123 Replicant Street, Los Angeles, California 90210--4321}}

\author{
    \IEEEauthorblockN{
        An-Dan Nguyen\IEEEauthorrefmark{1}, 
        Duc-Thinh Pham\IEEEauthorrefmark{2}, 
        Nimrod Lilith\IEEEauthorrefmark{1}, 
        and Sameer Alam\IEEEauthorrefmark{2}\IEEEauthorrefmark{1}
    }
        \IEEEauthorblockA{
        \IEEEauthorrefmark{1}Saab-NTU Joint Lab\\
}
    \IEEEauthorblockA{
        \IEEEauthorrefmark{2}Air Traffic Management Research Institute\\
        School of Mechanical \& Aerospace Engineering\\
        Nanyang Technological University, Singapore  \\       andan.nguyen@saabgroup.com, dtpham@ntu.edu.sg, nimrod.lilith@ntu.edu.sg, sameeralam@ntu.edu.sg}

}

\maketitle

\begin{abstract}
General real-time runway occupancy time prediction modelling for multiple airports is a current research gap. An attempt to generalize a real-time prediction model for Arrival Runway Occupancy Time (AROT) is presented in this paper by substituting categorical features by their numerical equivalences. Three days of data, collected from Saab Sensis' Aerobahn system at three US airports, has been used for this work. Three tree-based machine learning algorithms: Decision Tree, Random Forest and Gradient Boosting are used to assess the generalizability of the model using numerical equivalent features. We have shown that the model trained on numerical equivalent features not only have performances at least on par with models trained on categorical features but also can make predictions on unseen data from other airports.
\end{abstract}


\section{Introduction}
\label{sec:introduction-motivation}
Runways are one of the most valuable resources at an airport and one of the most important factors affecting airport capacity. (Arrival) Runway Occupancy Time (AROT) can be served as an indicator of runway efficiency \cite{spencer2019predictive}\cite{leighfisher2012evaluating}\cite{nikoleris2016effect}\cite{hu2019runway}. Minimizing the runway occupancy time has an operational impact, as stated in \cite{tether2003horndal} \cite{caa1993} as ``saving an average of 5 seconds on every aircraft's runway occupancy would add another 1 - 1.5 movements per hour [at London Heathrow]". A model which can predict the ROT accurately is, therefore, highly desirable. 

The availability of surveillance radar data has enabled ROT prediction and analysis research in recent years. Several studies have focused on making the model more general \cite{spencer2019predictive}, in order to be able to be applied to many different airports or to be a component in simulation software. Other works have attempted to make the real-time prediction \cite{friso2018predicting} \cite{dai2020runway} at different distances from the runway threshold. General prediction models which use data from various sources often have an advantage in terms of generalizability. However, the availability of the data sources must be satisfied.

On the other hand, real-time prediction models have often investigated the case of a specific airport. Consequently, the prediction at the airport can be very accurate. However, it may not be possible to apply to other airports due to the differences in data for each airport. In some instances, ROT data are not even available enough for retraining, especially at small and medium airports where traffic is relatively low.

\begin{figure*}
    \centering
    \begin{minipage}{0.495\textwidth}
        \centering
        \includegraphics[width=\textwidth]{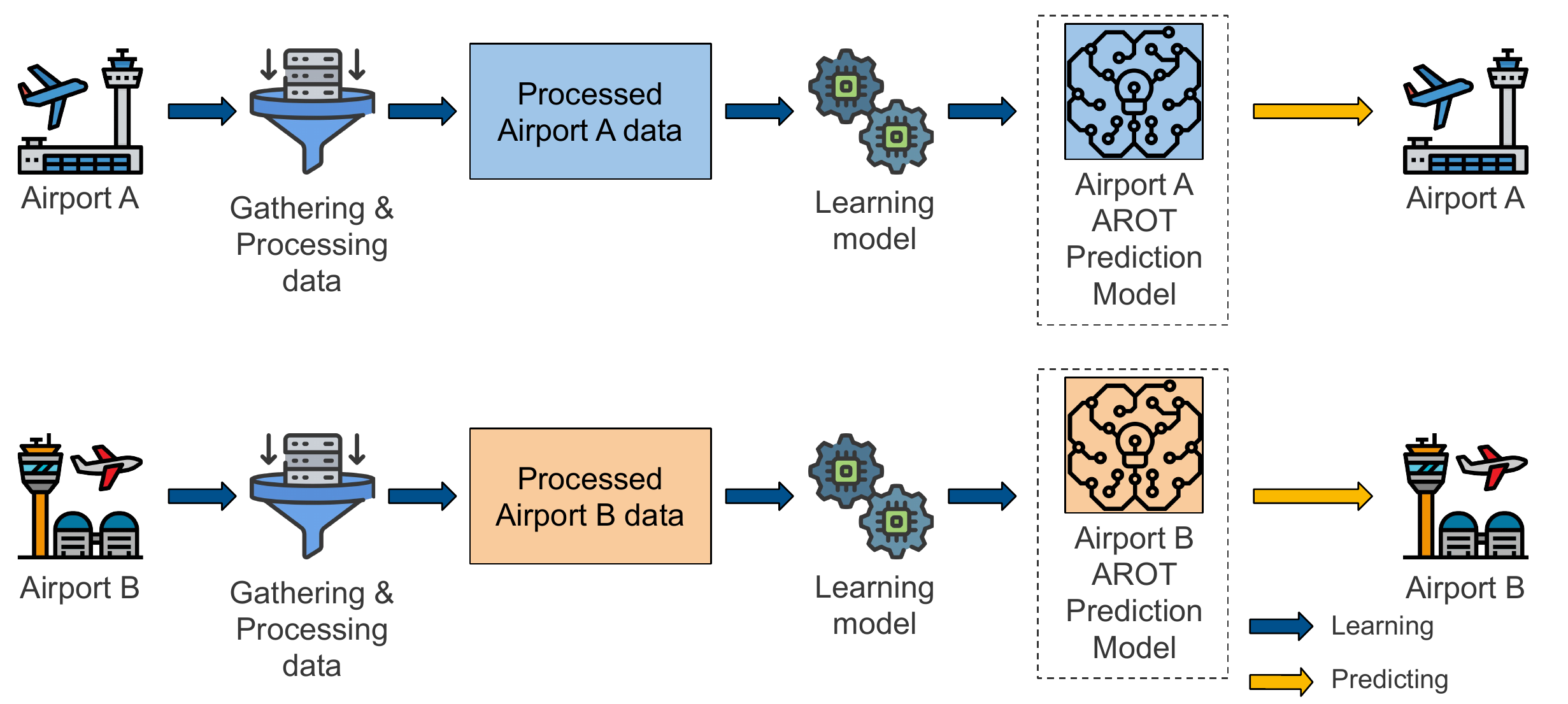}
        \caption{A normal way to make prediction model with separated data from two airports}
        \label{fig:normal-learning-process}
    \end{minipage}\hfill%
    \begin{minipage}{0.495\textwidth}
        \centering
        \includegraphics[width=\textwidth]{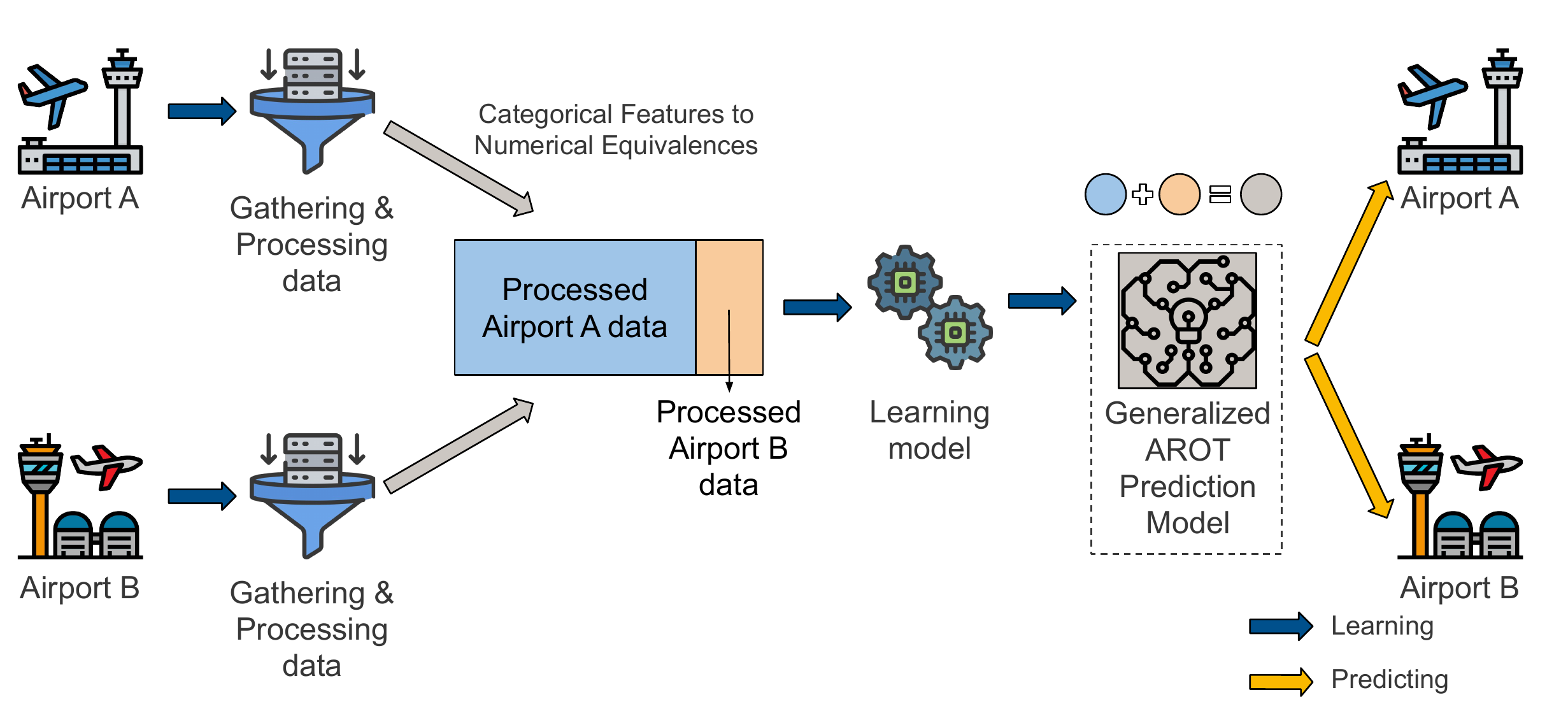}
        \caption{A way to make a generalized prediction model for two airport by transforming the dataset with feature equivalences}
        \label{fig:generalized-learning-process}
    \end{minipage}
\end{figure*}

In this research we present a way to have ROT prediction models train on some airports with limited data (3 days, around 1500 samples), and then be able to make predictions on other (target) airports by substituting categorical variables with their numerical equivalences without obtaining a large quantity of data from those target airports. This research can be summarized in Fig. \ref{fig:normal-learning-process} and Fig. \ref{fig:generalized-learning-process}. Fig. \ref{fig:normal-learning-process} depicts a conventional way of training two machine learning models on two different datasets from two airports. However, by substituting the categorical features with the numerical equivalences (will be described later in Section \ref{sec:feature-extraction/engineering}), as shown in Fig. \ref{fig:generalized-learning-process}, we create a generalized dataset that can serve as an input of machine learning algorithms to produce models which can predict AROT in both airport A and airport B. The benefit of the process is significant when airport B is a low traffic airport from which we cannot obtain a great deal of data.

The rest of this paper is organized as follows: we review some related works in Section \ref{sec:related-works}. In Section \ref{sec:data} and \ref{sec:feature-extraction/engineering} we give an overview of the data and the feature selection and engineering process. Section \ref{sec:pred-model} gives the details and the benefit of the prediction model. Section \ref{sec:experiments} details the way we conduct our experiments, which leads to the results in Section \ref{sec:results}. We gave our final conclusions in Section \ref{sec:conclusion}.

\section{Related Works}
\label{sec:related-works}
Runway occupancy time (ROT) has long been an important part in research about runway operation and runway efficiency. One of the early investigations on ROT was the work of Koenig \cite{koenig1978analysis}. The paper used data collected from six airports in the US between 1972 to 1973 to identify the patterns that contribute to high ROT, which in turn causes delay and reduces airport capacity. The author identified the possible influential factors of ROT, such as company procedure, gate location, pilot's knowledge of a specific runway and the design of the runways and runway exits. Some factors are easier to identify than others. Nevertheless, the author, through analysis, concluded that the most important factor is the position of the runway exit taken relative to the terminal gate.

Following the research of Keonig, Weiss and Barrer \cite{weiss1984analysis} conducted an analysis of ROT and the relation of longitudinal separation using data collected at three airports: LaGuardia airport, Newark airport and Boston Logan International airport. The data were collected through visual means at the airport tower and, in some cases, nearby locations where all runways and exits were visible. Although there were some difficulties in data collection, since the threshold crossing time and time clear of the runway had to be estimated visually, nearly 600 data points were collected at each airport. After analyzing the data, the author concluded that the condition of the runway (wet/dry) did not have a significant effect on the difference of average ROTs.

As the data became more abundant and more accessible, thanks to technologies such as Automatic Dependent Surveillance—Broadcast (ADS-B) and Advanced Surface Movement Guidance \& Control System (A-SMGCS), runway occupancy time research using data-driven methods has become increasingly active in recent years. One of the demonstrations of this technique, using surveillance data in measuring ROT, was the work of Kumar et al.\cite{kumar2009runway}. In this study, the authors presented an algorithm to measure the ROT using three days' radar track data extracted from Airport Surface Detection Equipment, Model X (ASDE-X) system. The subsequent analysis on the extract ROTs showed that small aircraft have ROTs as high as large aircraft. Furthermore, identical runways, which have the same measurement in length and width, can exhibit significantly different ROTs.

Kolos-Lakatos\cite{kolos2013influence} also used data from the ASDE-X system to assess the influence of two major elements on runway capacity, that is ROT and wake vortex separation. The data were collected at Boston, Philadelphia, New York La Guardia and Newark airports. The authors found that the size of aircraft is not a strong factor affecting ROT, as small aircraft occupy the runway as long as the large aircraft. The same applies for runway occupancy times in Visual Meteorological Conditions (VMC) and Instrument Meteorological Conditions (IMC). However, runways equipped with high-speed exits showed a reduction in ROT compared to runway equipped with standard 90-degree exits.

Recently, machine learning techniques have been applied in predicting ROT and some studies have shown promising results. Martinez et al.\cite{martinez2018boosted} applied Gradient Boosting Tree Framework to predict, at various distances from runway threshold, the runway exit a flight will take and predicted the arrival ROTs at 2 nautical miles (NM) from runway threshold. The authors managed to get an Area Under the Curve (AUC) of 0.784 in predicting runway exits and a mean absolute error (MAE) of 8 seconds compared to a standard deviation of 14.4 seconds in ROT predictions.

In another work Friso et al.\cite{friso2018predicting} did not make a generic ROT prediction model. Instead, the authors focused on the detection of abnormal AROTs and used a Decision Tree to build the ``what-if" statements. And, from these results, the authors identified 17 abnormal AROT categories and their related precursors. Moreover, the authors also built a real time model to predict abnormal AROTs.

Spencer and Trani\cite{spencer2019predictive} have taken a different approach. Instead of making a model specific to one airport, the authors used statistical modeling and data from 27 airports with over 800,000 data points to model departure and arrival ROT. One of the advantages of the authors' model is the generalizability due to the choice of a small set of general features. However, the lack of real-time aircraft features and the assumption of runway exit information may inhibit the capability of applying this model in a real-time prediction.

Meijers\cite{meijers2020data} recently presented a comprehensive work on the analysis and prediction of ROT using data-driven methods. The work detailed the way to collect ROT data, identify factors affecting ROT and presented a two-step model to predict ROT using the data from 36 airports. The two-step model utilized variants of Neural Network to predict the ROT distribution of landing flight. Hence, there is a lack of interpretability and accumulation of errors in this approach. The author also presented a case study at Keflavík airport where the predictive models suggested locations of new exits which had the potential of reducing the average ROT by up to 23 seconds during peak-hour.

A recent study of Dai and Hansen\cite{dai2020runway} presented a real-time prediction of Runway Occupancy Buffers (ROBs), a new concept proposed by the authors to measure the difference in time when the leading aircraft exits the runway and the trailing aircraft crosses the runway threshold. The authors used several machine learning techniques to predict the ROBs at different distances from the runway thresholds. The best model achieved an R-squared of up to 90\%. A limitation, and an extension of this work, stated by the authors, is applying the model at different airports.

This study is an attempt to combine the benefits of a real-time prediction model and a generalized one, which is also currently a gap in ROT studies. We present a real-time prediction model that can give ATCO arrival runway occupancy time information 75 to 120 seconds before the aircraft crosses the runway threshold. Moreover, the prediction model can also be applied to other airports, which is our focus on this study by applying our pre-processing techniques of substituting categorical features by numerical equivalences.

\section{Data}
\label{sec:data}
The prediction framework makes use of three data categories:
\subsection{Flight data}
The raw flight data are collected using Saab Sensis' Aerobahn system from three airports: 
\begin{itemize}
    \item Ronald Reagan Washington National Airport (ICAO: KDCA, IATA: DCA), which has 1232 samples.
    \item Miami International Airport (ICAO: KMIA, IATA: MIA), which has 1628 samples.
    \item Phoenix Sky Harbor International Airport (ICAO: KPHX, IATA: PHX), which has 1722 samples
\end{itemize}
All sets of data range from 16.05.2019 to 18.05.2019 inclusive, that is, three days. The raw data come from two different sources:
\begin{itemize}
    \item The Region Occupancy Reports, which record how long an airplane occupies the regions surrounding and at the airport such as: terminal maneuvering area (TMA), runway, taxiway, ramp and gate. The Region Occupancy Reports also include the information related to a flight, such as origin, destination, airline, aircraft type, gate assigned and runway assigned.
    \item The Surveillance Track Data, which records the position, speed, flight level and heading of an aircraft at a specific point in time.
\end{itemize}
We summarize the data components from these two sources in two tables, Table \ref{tab:region-occupancy-data} and Table \ref{tab:surveillance-track-data}, below.
\begin{table}[htbp]
\caption{Descriptions of data from Region Occupancy Reports}
\centering
\begin{tabularx}{\columnwidth}{|p{2cm}|X|l|}
\hline
\textbf{Name}            & \textbf{Description}                                                    & \textbf{Data type} \\ \hline
Call sign                & The call sign of the flight                                             & Categorical        \\ \hline
Airline                  & The airline that operate the flight                                     & Categorical        \\ \hline
Aircraft type            & Type of the aircraft                                                    & Categorical        \\ \hline
Maximum landing weight   & The maximum landing weight of the aircraft                              & Numerical          \\ \hline
Origination airport code & The origin airport of the flight                                        & Categorical        \\ \hline
Destination airport code & The destination airport of the flight                                   & Categorical        \\ \hline
Flight original date     & The date when the flight depart                                         & Datetime           \\ \hline
Gate assigned            & Arrival gate assigned to the flight                                     & Categorical        \\ \hline
Runway assigned          & Runway the flight used to land                                          & Categorical        \\ \hline
Region occupied name     & The name of the region at the airport defined by the information system & Categorical        \\ \hline
Region type              & The type of the region. It can be: TMA, Runway, Taxiway, Ramp, Gate     & Categorical        \\ \hline
Time entered             & Timestamp when the aircraft enters the region                           & Datetime           \\ \hline
Time exited              & Timestamp when the aircraft completely vacates the region               & Datetime           \\ \hline
Occupancy time           & The time the aircraft occupies the region                               & Numerical          \\ \hline
\end{tabularx}
\label{tab:region-occupancy-data}
\end{table}

\begin{table}[htbp]
\caption{Descriptions of Surveillance Track Data}
\centering
\begin{tabularx}{\columnwidth}{|p{2cm}|X|l|}
\hline
\textbf{Name}            & \textbf{Description}                                  & \textbf{Data type} \\ \hline
Timestamp                & Time when the data is recorded                        & Datetime           \\ \hline
Call sign                & Call sign of the flight                               & Categorical        \\ \hline
Track position           & Position of the flight in X/Y/Z Cartesian coordinates & Numerical          \\ \hline
Origination airport code & The origin airport of the flight                      & Categorical        \\ \hline
Destination airport code & The destination airport of the flight                 & Categorical        \\ \hline
Measured flight level    & Flight level of the aircraft                          & Numerical          \\ \hline
Track heading            & The heading of the aircraft                           & Numerical          \\ \hline
Speed                    & The true air speed of the aircraft                    & Numerical          \\ \hline
\end{tabularx}
\label{tab:surveillance-track-data}
\end{table}

We combine data from these two sources by matching call sign, origination airport code, destination airport code and flight original date, to gather all the information of a given flight for feature extraction and engineering. We also extract the recorded arrival runway occupancy time through Region Occupancy Reports to serve as a target for the prediction model. We will present these in detail in  Section \ref{sec:feature-extraction/engineering}.

\subsection{Weather data}
We collect the weather data from Surface Weather Observation Station (ASOS/AWOS) at the three airports mentioned above during the same three days via \href{https://mesonet.agron.iastate.edu/request/download.phtml}{\color{blue}Iowa Environmental Measonet (IEM)}. The weather data fields are shown in Table \ref{tab:weather-data}.
\begin{table}[htbp]
\caption{Descriptions of weather data}
\centering
\begin{tabularx}{\columnwidth}{|p{2cm}|X|l|}
\hline
\textbf{Name}     & \textbf{Description}                    & \textbf{Data type} \\ \hline
Timestamp         & Time when the weather data are recorded & Datetime           \\ \hline
Temperature       & Air temperature in Fahrenheit           & Numerical          \\ \hline
Visibility        & Visibility in miles                     & Numerical          \\ \hline
Wind direction    & Wind direction in degrees from north    & Numerical          \\ \hline
Wind speed        & Wind speed in knots                     & Numerical          \\ \hline
Pressure altimeter & Pressure altimeter in inches           & Numerical          \\ \hline
\end{tabularx}
\label{tab:weather-data}
\end{table}

\subsection{Supplemental Data}
Furthermore, we also collected additional data as described in Table \ref{tab:supplemental-data} about runways at each airport, using information from the website  \href{https://airnav.com/}{\color{blue}AirNav.com}. The data are measurements of runways such as runway length, runway width, runway altitude and runway true heading. We collected this data as a substitution for the runway assigned feature, which we will explain in detail in Section \ref{sec:feature-extraction/engineering}.
\begin{table}[htbp]
\caption{Descriptions of Supplemental Data}
\centering
\begin{tabularx}{\columnwidth}{|p{2cm}|X|l|}
\hline
\textbf{Name}       & \textbf{Description}                  & \textbf{Data type} \\ \hline
Runway length       & Length of the runway                  & Numerical          \\ \hline
Runway width        & Width of the runway                   & Numerical          \\ \hline
Runway altitude     & Altitude of the runway                & Numerical          \\ \hline
Runway true heading & Heading of the runway from true north & Numerical          \\ \hline
\end{tabularx}
\label{tab:supplemental-data}
\end{table}

\section{Feature Extraction and Engineering}
\label{sec:feature-extraction/engineering}
Using the data sources above, and getting inspiration from some of the past works \cite{spencer2019predictive}\cite{friso2018predicting}\cite{dai2020runway}\cite{martinez2018boosted}, especially the data-driven analysis of Meijers and Hansman\cite{meijers2019data}, we select some of the fields to be the features we use to train our machine learning models. We did not collect and include data about runway exit locations and types, as we are making a real-time AROT prediction model around the final approach fix where the runway exit locations and types information would not be yet known.

The features of the model to predict AROT can be divided into five main groups:
\begin{itemize}
    \item \textbf{Airport features} which is the information of the runway a given aircraft used to land, such as runway (name) assigned, runway length, runway width, runway altitude, runway true heading and the gate assigned to the flight. The runway length, runway width, runway altitude and runway true heading are the numerical equivalences of the runway assigned feature. We also extract the distance from the last trajectory point, when the aircraft fully stops, to the runway it used to land. We use this numerical feature as a substitute for the assigned gate feature, which is categorical.
    \item \textbf{Aircraft features} which are the aircraft's type and maximum landing weight. The maximum landing weight is the numerical equivalence of the aircraft type categorical feature.
    \item \textbf{Aircraft information at/near final approach fix features} which are the distance to the runway threshold, the flight level and the true heading of the aircraft at the time of final approach fix. This information is extracted based on the Surveillance Track Data described above.
    \item \textbf{Weather information features} which are temperature, visibility, wind direction, wind speed and pressure altimeter.
    \item \textbf{Short term runway usage statistics features} which are number of aircraft landing in the last 30 minutes of a specific runway and the last 30 minutes' average AROT.
\end{itemize}

The features of the model are shown in Table \ref{tab:features} below.
\begin{table}[htbp]
\caption{Input Features to Machine Learning Model}
\begin{tabularx}{\columnwidth}{|p{2cm}|X|l|}
\hline
\textbf{Name}                                 & \textbf{Description}                                                                           & \textbf{Data type} \\ \hline
Runway assigned                               & Runway the flight used to land                                                                 & Categorical        \\ \hline
Runway length                                 & Length of the runway                                                                           & Numerical          \\ \hline
Runway width                                  & Width of the runway                                                                            & Numerical          \\ \hline
Runway altitude                               & Altitude of the runway                                                                         & Numerical          \\ \hline
Runway true heading                           & Heading of the runway from true north                                                          & Numerical          \\ \hline
Gate assigned                                 & Arrival gate assigned to the flight                                                            & Categorical        \\ \hline
Distance from last point trajectory to runway & The distance from the last point of the trajectory when the aircraft fully stops to the runway & Numerical          \\ \hline
Aircraft type                                 & Type of the aircraft                                                                           & Categorical        \\ \hline
Maximum landing weight                        & The maximum landing weight of the aircraft                                                     & Numerical          \\ \hline
Distance to the runway threshold              & Distance to of the aircraft to the runway threshold at/near final approach fix                 & Numerical          \\ \hline
Flight level                                  & Flight level of aircraft at/near final approach fix                                            & Numerical          \\ \hline
True heading                                  & True heading of aircraft at/near final approach fix                                            & Numerical          \\ \hline
Temperature                                   & Air temperature in Fahrenheit                                                                  & Numerical          \\ \hline
Visibility                                    & Visibility in miles                                                                            & Numerical          \\ \hline
Wind direction                                & Wind direction in degrees from north                                                           & Numerical          \\ \hline
Wind speed                                    & Wind speed in knots                                                                            & Numerical          \\ \hline
Pressure altimeter                            & Pressure altimeter in inches                                                                   & Numerical          \\ \hline
Number of flight landed last 30 minutes       & Number of flight landed on the runway in the last 30 minutes                                   & Numerical          \\ \hline
Average AROT last 30 minutes                  & Average AROT of flight landed on the runway in the last 30 minutes                             & Numerical          \\ \hline
\end{tabularx}
\label{tab:features}
\end{table}

The target of prediction is the Arrival Runway Occupancy Time (AROT) at each airport. The histograms of the AROTs at those airports are presented below.
\begin{figure}[htbp]
\centerline{\includegraphics[width=0.9\columnwidth]{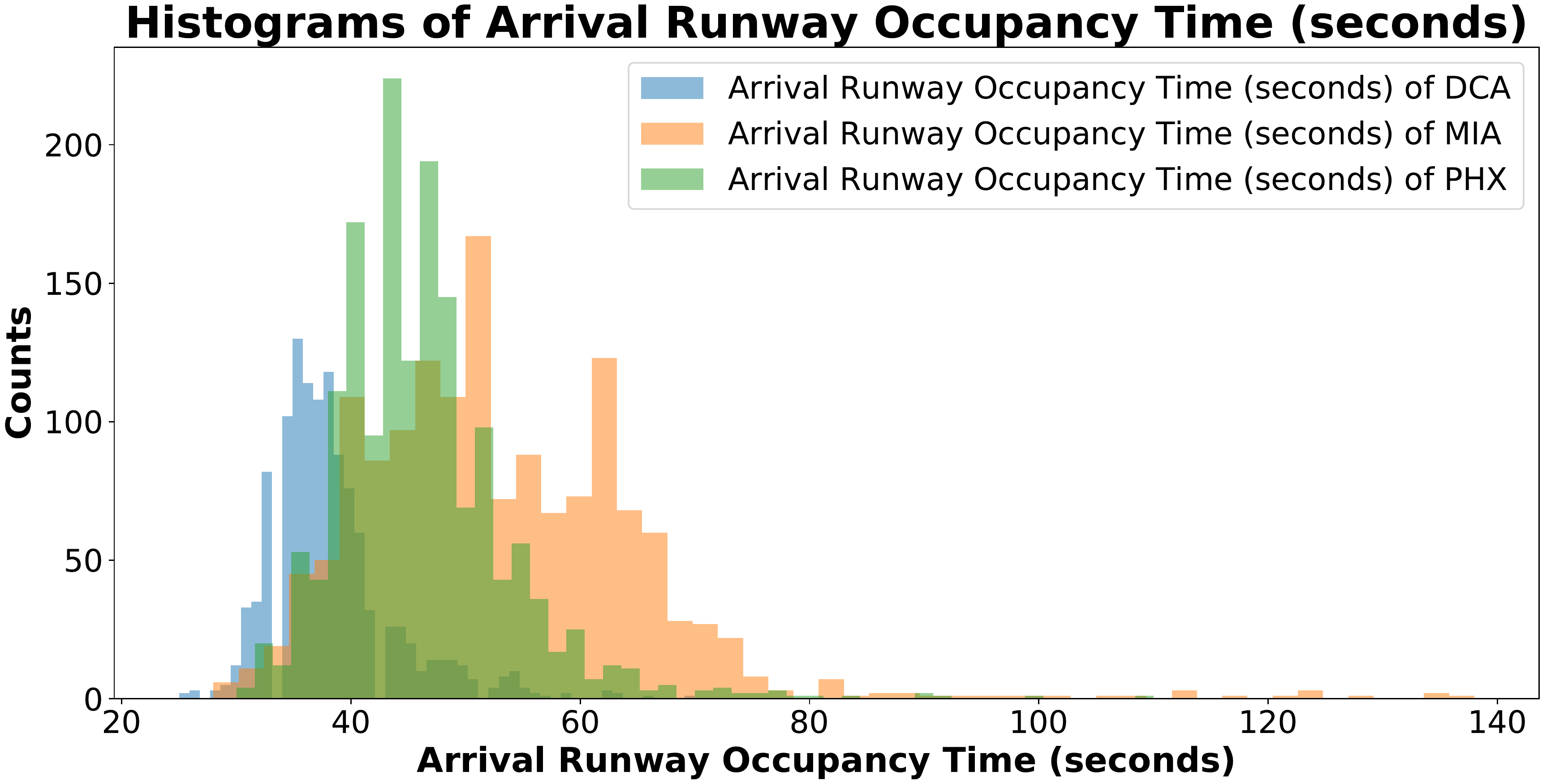}}
\caption{Histogram of Arrival Runway Occupancy Time at three airports: DCA, MIA, PHX}
\label{fig:arot-hist}
\end{figure}

As we mentioned in the Section \ref{sec:introduction-motivation}, one of the advantages of our method is the AROT prediction model can be applied, that is make AROT predictions, for other airports without having to retrain the model with the data at those airports. This generalizability of the prediction model is particularly useful in the case where the AROT data is limited or not available, for example, in small and medium airports where traffic is low. We enable the generalizability by substituting the categorical features of the training data by their numerical equivalences. 

Although categorical features are easier for a human to understand, extra processing steps, such as ordinal-encoding, one-hot-encoding, or hashing, are needed to make the data trainable in almost all machine learning algorithms. These extra steps can be eliminated by substituting categorical features by their numerical equivalences. Moreover, methods such as one-hot-encoding can make the post-processed features space increase up to the sum of values in those categorical features. This makes the model more prone to the curse of dimensionality, which is undesirable. We present these categorical features and their numerical equivalences in Table \ref{tab:feature-equivalences} as follows.
\begin{table}[htbp]
\caption{Categorical Features and Numerical Equivalences}
\begin{tabularx}{\columnwidth}{|p{2.5cm}|X|}
\hline
\textbf{Categorical features}    & \textbf{Numerical equivalences}               \\ \hline
Aircraft type                    & Maximum landing weight                        \\ \hline
\multirow{4}{*}{Runway assigned} & Runway length                                 \\ \cline{2-2} 
                                 & Runway width                                  \\ \cline{2-2} 
                                 & Runway altitude                               \\ \cline{2-2} 
                                 & Runway true heading                           \\ \hline
Gate assigned                    & Distance from last point trajectory to runway \\ \hline
\end{tabularx}
\label{tab:feature-equivalences}
\end{table}

In this work we have three different datasets: categorical, numerical and mixed dataset. A categorical dataset is a dataset that has only categorical features, but not their numerical equivalences. A numerical dataset is a dataset has the numerical equivalent features, but not their categorical counterpart. And a mixed dataset is a dataset that has all the features mentioned above, that is, including both the categorical features and their numerical equivalences. From this point on we call a prediction model trained on the categorical dataset a categorical model. Likewise, we a numerical model and a mixed model are trained on a numerical dataset and a mixed dataset respectively. The details of the AROT prediction model, and its benefits, are presented in Section \ref{sec:pred-model}.

\section{AROT Prediction Model}
\label{sec:pred-model}
We train the prediction model utilizing the five groups features shown in Table \ref{tab:features}. As defined in the previous section, we have three types of model: categorical model, numerical model and mixed model. We used a set of features to train a corresponding model, e.g. categorical features to train categorical models using three learning algorithms: Decision Tree Regressor, Random Forest Regressor, and Gradient Boosting Regressor with implementation from Scikit-learn \cite{scikit-learn}. We also employed grid-search (details in Table \ref{tab:hyper-param-ranges}) with the implementation from Scikit-learn for hyper-parameters tuning for each learning algorithm. We present an experiment to compare the performance of these models and show the results in Sections \ref{sec:experiments} and \ref{sec:results}.

One of the main benefits of our model is to be able to make a real-time AROT prediction at, or near, the final approach fix of a runway at an airport. Table \ref{tab:avg-pred-point-measurement} shows the average distance and time for the aircraft to reach the runway threshold at each runway at each airport when the prediction happens. The information is extracted from our dataset. The column ``Average Seconds to Threshold" is the average amount of time in seconds prior to runway threshold crossing that the model can provide an AROT prediction to the ATCO. In other words, using this model, the ATCO can be aware of the AROT 75 to 120 seconds before the aircraft crosses the runway threshold. This can be a valuable time for the ATCO to make a crucial decision if needed. Future work will try to further extend the prediction point beyond the final approach fix.
\begin{table}[htbp]
\caption{Prediction point Average Distance and Time to Threshold}
\label{tab:avg-pred-point-measurement}
\begin{tabularx}{\columnwidth}{|Y|Y|Y|Y|Y|}
\hline
\textbf{Airport} & \textbf{Runway Name} & \textbf{Average Distance from Threshold (NM)} & \textbf{Average Speed (knot)} & \textbf{Average Seconds to Threshold (second)} \\ \hline
\multirow{3}{*}{DCA} & 1   & 4.82 & 166.38 & 105.05 \\ \cline{2-5} 
                     & 19  & 6.18 & 179.82 & 124.62 \\ \cline{2-5} 
                     & 33  & 3.42 & 163.96 & 75.63  \\ \hline
\multirow{4}{*}{MIA} & 8L  & 4.45 & 157.96 & 102.29 \\ \cline{2-5} 
                     & 8R  & 4.45 & 151.81 & 106.61 \\ \cline{2-5} 
                     & 9   & 4.29 & 162.61 & 95.79  \\ \cline{2-5} 
                     & 12  & 5.97 & 177.9  & 121.96 \\ \hline
\multirow{5}{*}{PHX} & 7R  & 5.75 & 186.15 & 112.08 \\ \cline{2-5} 
                     & 8   & 5.51 & 182.63 & 109.2  \\ \cline{2-5} 
                     & 25L & 5.63 & 173.17 & 118.05 \\ \cline{2-5} 
                     & 25R & 5.56 & 170.78 & 118.37 \\ \cline{2-5} 
                     & 26  & 5.6  & 173.01 & 117.47 \\ \hline
\end{tabularx}
\end{table}

In Section \ref{sec:experiments}, we detail the experiments where we compare the performance of the  categorical, numerical and mixed models, and their generalized learning results. We also specify the experiment where we show the generalizability and the benefit of the numerical models.

\section{Experiments}
\label{sec:experiments}
\subsection{Arrival Runway Occupancy Time (AROT) Learning Experiment}
In this experiment we trained three machine learning algorithms mentioned above for three types of dataset (numerical, categorical and mixed) in three different airports (DCA, MIA and PHX) we mentioned in the previous sections.

% The overview of the learning experiment process is shown in Fig. \ref{fig:learning-experiment}.
% \begin{figure}[htbp]
% \centerline{\includegraphics[width=0.95\columnwidth]{figure/learning-experiment-v1-1.pdf}}
% \caption{Overview of AROT Learning Experiment.}
% \label{fig:learning-experiment}
% \end{figure}

We employ the cross-validation technique to assess the capability of prediction of unseen data of these models. As the training folds are changed during cross-validation, the hyper-parameters of those learning algorithms must be changed accordingly. In order to obtain the best hyper-parameters for the learning algorithm, we utilize the cross-validation technique again on those training folds. The hyper-parameter set that has the lowest error according to the cross-validation results is used to refit the data in training folds then make predictions on the testing fold.  Table \ref{tab:hyper-param-ranges} shows the hyper-parameters search ranges of the learning algorithms.
\begin{table}[htbp]
\centering
\caption{The search range of learning algorithm's hyper-parameters}
\begin{tabularx}{\columnwidth}{|X|Y|Y|Y|}
\hline
\textbf{Hyper-parameters} & \textbf{Range in Decision Tree Regressor} & \textbf{Range in Random Forest Regressor} & \textbf{Range in Gradient Boosting Regressor} \\ \hline
splitter           & [best, \textbf{random}]                 & N/A                                  & N/A                                         \\ \hline
max\_depth         & [\textbf{3}, \textbf{10}, 30, 100, 300] & [3, \textbf{10}, 30, 100]            & [\textbf{3}, \textbf{10}, \textbf{30}, 100] \\ \hline
min\_sample\_leaf  & [1, 5, \textbf{10}, \textbf{30}]        & [1, \textbf{5}, \textbf{10}, 30]     & [1, \textbf{5}, \textbf{10}, \textbf{30}]   \\ \hline
max\_features      & [sqrt, 10, \textbf{auto}]               & [\textbf{sqrt}, 10, auto]            & [\textbf{sqrt}, \textbf{10}, auto]          \\ \hline
n\_estimators      & N/A                                     & [10, 30, \textbf{100}, \textbf{300}] & [30, 100, \textbf{300}, 900]                \\ \hline
max\_samples       & N/A                                     & [0.1, 0.3, \textbf{None}]            & N/A                                         \\ \hline
learning\_rate     & N/A                                     & N/A                                  & [0.001, 0.003, \textbf{0.01}, 0.03, 0.1]    \\ \hline
subsample          & N/A                                     & N/A                                  & [0.1, \textbf{0.3}, \textbf{1.0}]           \\ \hline
\end{tabularx}
\label{tab:hyper-param-ranges}
\end{table}

We highlighted the best settings of each algorithm in the table above. The best settings showed an inclination towards a generalized model in each algorithm. For example, a shallow tree which has leaves contain a large number of nodes has more generalization capacity than a deep tree which has leaves contains a small number of nodes. Because shallow tree, which has large a number of samples in leaves node, will avoid going into particular cases when making predictions. Our learning algorithm best model setting prefer the shallow tree (small value of \textit{max\_depth} resulted in best models more often than large value). The same conclusion can be drawn when looking at setting \textit{min\_sample\_leaf} when larger values are preferred.

\subsection{Generalized Learning Experiment}
\begin{figure}[htbp]
\centerline{\includegraphics[width=0.935\columnwidth]{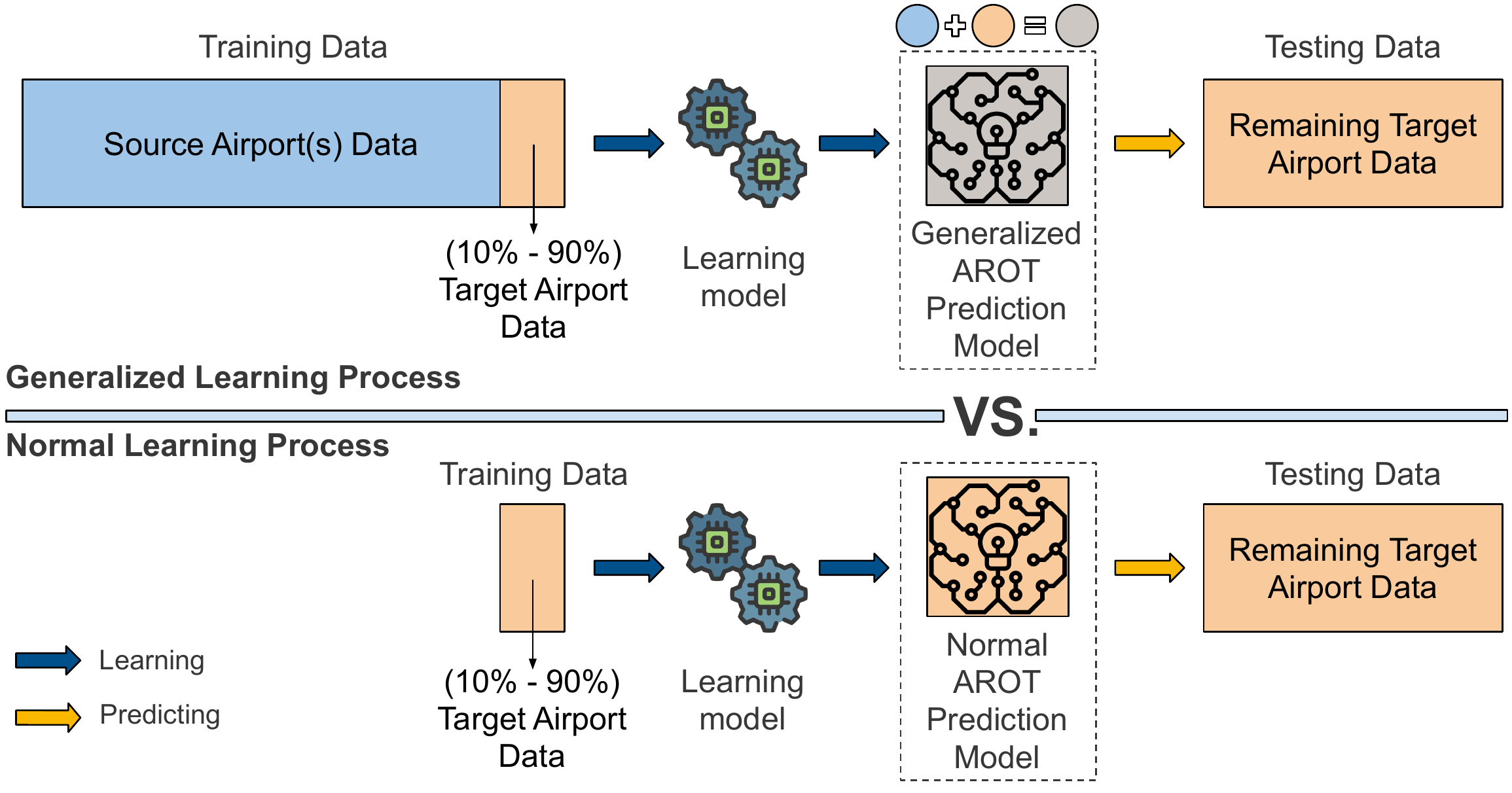}}
\caption{Overview of Generalized Learning Experiment}
\label{fig:generalized-learning-experiment}
\end{figure}

In order to evaluate the generalizability of the prediction model trained on the numerical dataset, we design an experiment where we train the prediction model using the three machine learning algorithms mentioned above, on the source airport(s) numerical data together with a fraction ($\alpha = 0.1 \dots 0.9$) of the target airport data to make AROT predictions on target airport. This model is called the Generalized Model. We also train a prediction model only with a fraction ($\alpha = 0.1 \dots 0.9$) of target airport data with no source airport data to compare with the generalized learning model. This model is called the Normal Model. The experiment is illustrated in Fig. \ref{fig:generalized-learning-experiment}. This figure is the detailed version of Fig.  \ref{fig:normal-learning-process} and Fig. \ref{fig:generalized-learning-process}, when we realize the idea presented in those two previous figures into a measurable experiment.

We have two versions of this generalized learning experiment. In the first version, the Generalized Model uses the data from a single source airport along with a fraction of the data from the target airport and obtains the prediction errors on the remainder of the target data.  In the second version, the Generalized Model uses the data from two source airports along with a fraction of data from the target airport and obtain the prediction errors on the remainder of the target data. We present the results of the two learning experiments in the next Section \ref{sec:results}.

\section{Results}
\label{sec:results}
\subsection{Unseen Data Prediction Results}
\begin{figure*}[htbp]
\centerline{\includegraphics[width=0.613\textwidth]{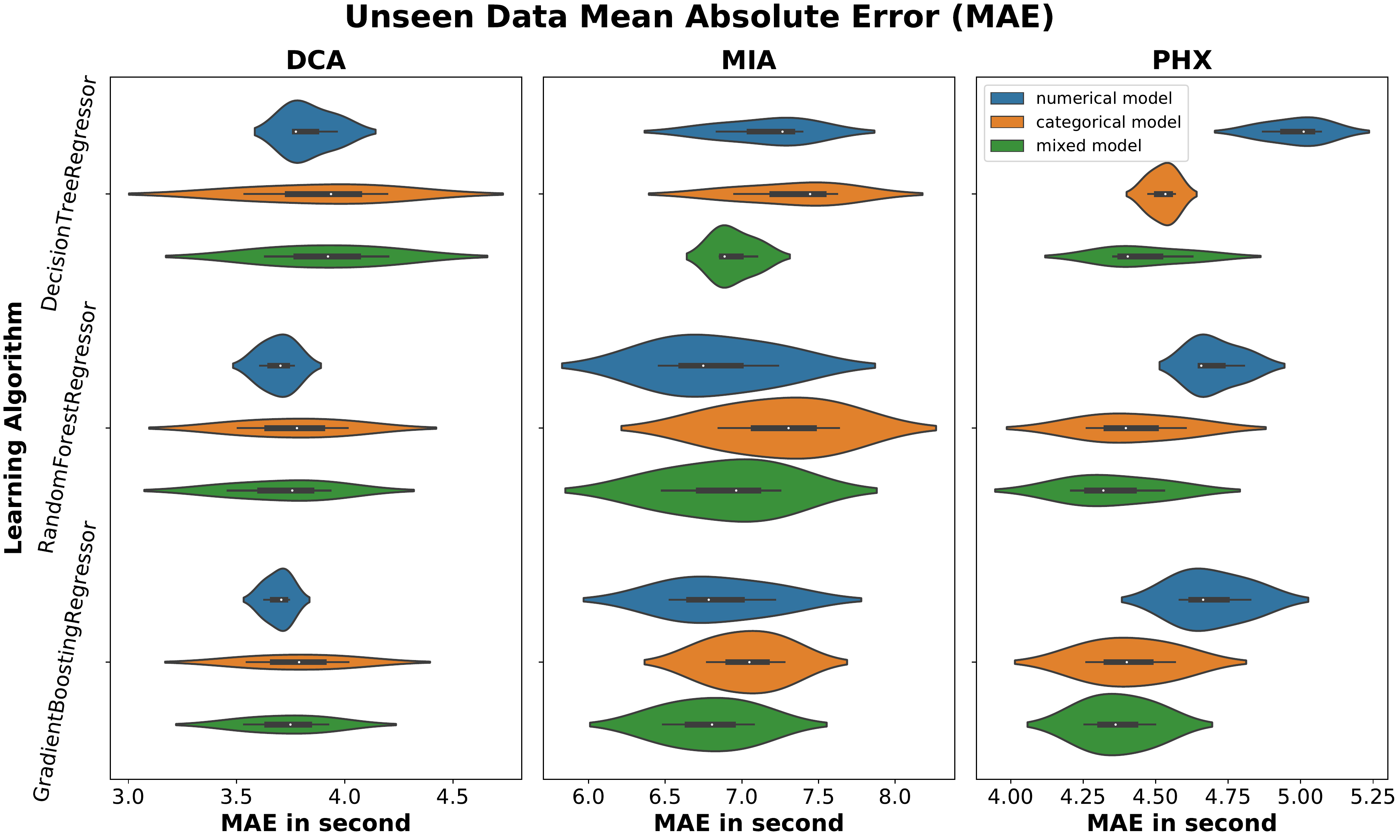}}
\caption{Results of three dataset (DCA, MIA, and PHX) are shown in three vertical plots (stacked horizontally). Horizontal axis is Mean Absolute Error in second. Vertical axis is different machine learning algorithms apply on the datasets. Dataset type (numerical, categorical, and mixed) is color coded.}
\label{fig:unseen-data-performances}
\end{figure*}

As shown in Fig. \ref{fig:unseen-data-performances}, we can see the numerical models are better than the categorical and mixed models in six out of nine cases (data from DCA and MIA) in terms of median of the predictions. However, among those six cases, the variances of the predictions of the numerical models are better than the others only for the DCA data.

At Phoenix Sky Harbor International Airport (PHX), the categorical models outperformed the numerical models. Moreover, the mixed models' performance (however, not clearly) exceeds both the numerical and categorical models. On one hand, this may suggest that the categorical and numerical equivalent features alone are not adequate for the prediction models in this particular case. On the other hand, this also may suggest that there is a multicollinearity effect in play in this circumstance, as categorical features and their numerical equivalences are just different representations of the same concepts. This phenomenon will be investigated further in upcoming work.

We present the mean and standard deviation AROT at each airport to compare with the Gradient Boosting Regressor prediction mean absolute errors (MAE) in Table \ref{tab:gbr-pred-error} below. Based on the model mean absolute error, we can see the learned model often reduces the uncertainty of the target airport AROT by 32\%-47\%. The prediction error can possibly be greatly reduced by the use of more data than three days data we have used in this work.

\begin{table}[htbp]
\caption{Gradient Boosting Regressor Prediction Error and AROT Range}
\label{tab:gbr-pred-error}
\begin{tabularx}{\columnwidth}{|Y|Y|Y|Y|}
\hline
\textbf{Airport}     & \textbf{AROT Range (second)}                            & \textbf{Model Type} & \textbf{Model's MAE (second)} \\ \hline
\multirow{3}{*}{DCA} & \multirow{3}{*}{$38.15 \pm 5.46$} & numerical           & $3.69 \pm 0.06$                               \\ \cline{3-4} 
                     &                              & categorical & $3.78 \pm 0.24$ \\ \cline{3-4} 
                     &                              & mixed       & $3.74 \pm 0.2$  \\ \hline
\multirow{3}{*}{MIA} & \multirow{3}{*}{$52.88 \pm 12.65$} & numerical   & $6.84 \pm 0.35$ \\ \cline{3-4} 
                     &                              & categorical & $7.03 \pm 0.25$ \\ \cline{3-4} 
                     &                              & mixed       & $6.79 \pm 0.3$  \\ \hline
\multirow{3}{*}{PHX} & \multirow{3}{*}{$46.08 \pm 7.57$}  & numerical   & $4.69 \pm 0.12$ \\ \cline{3-4} 
                     &                              & categorical & $4.41 \pm 0.15$ \\ \cline{3-4} 
                     &                              & mixed       & $4.37 \pm 0.12$ \\ \hline
\end{tabularx}
\end{table}

\subsection{Generalized Learning Results}
\begin{figure*}[htbp]
    \centering
    \includegraphics[width=0.613\textwidth]{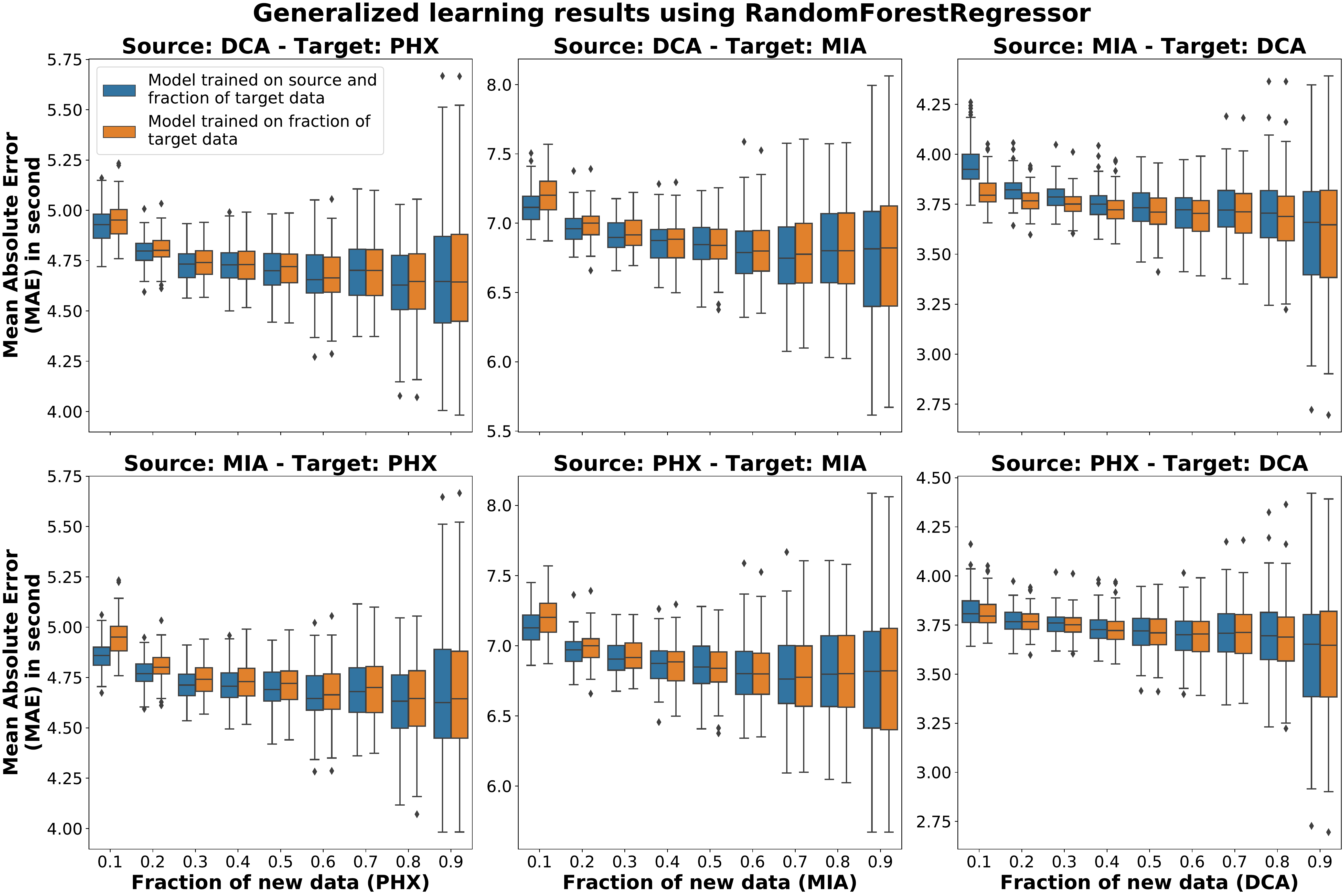}
    \caption{Generalized learning results using Random Forest Regressor. The train data is from source airport and a fraction ($\alpha = 0.1 \dots 0.9$) of target airport data. Testing is performed on the remainder ($1 - \alpha$) of target airport data. Evaluation is done using Mean Absolute Error in second.}
    \label{fig:generalized-learning-one-source-algo-rf}
\end{figure*}

In the first version of the generalized learning experiment, we train the Generalized Model on data from only one source airport along with a fraction of data from the target airport, then use that model to make predictions on the remaining target data. The Random Forest Regressor model performs the best, the generalized learning benefits being evident as the error ranges and medians of the Generalized Model are smaller than the Normal Model, when the fraction of target data is small ($\alpha = 0.1 \dots 0.4$) in 4 out of the 6 cases we consider, as shown in Fig. \ref{fig:generalized-learning-one-source-algo-rf}. When the fraction of target data is large ($\alpha = 0.5 \dots 0.9$), the generalized benefits are not very clear.

\begin{figure*}[htbp]
    \centering
    \includegraphics[width=0.613\textwidth]{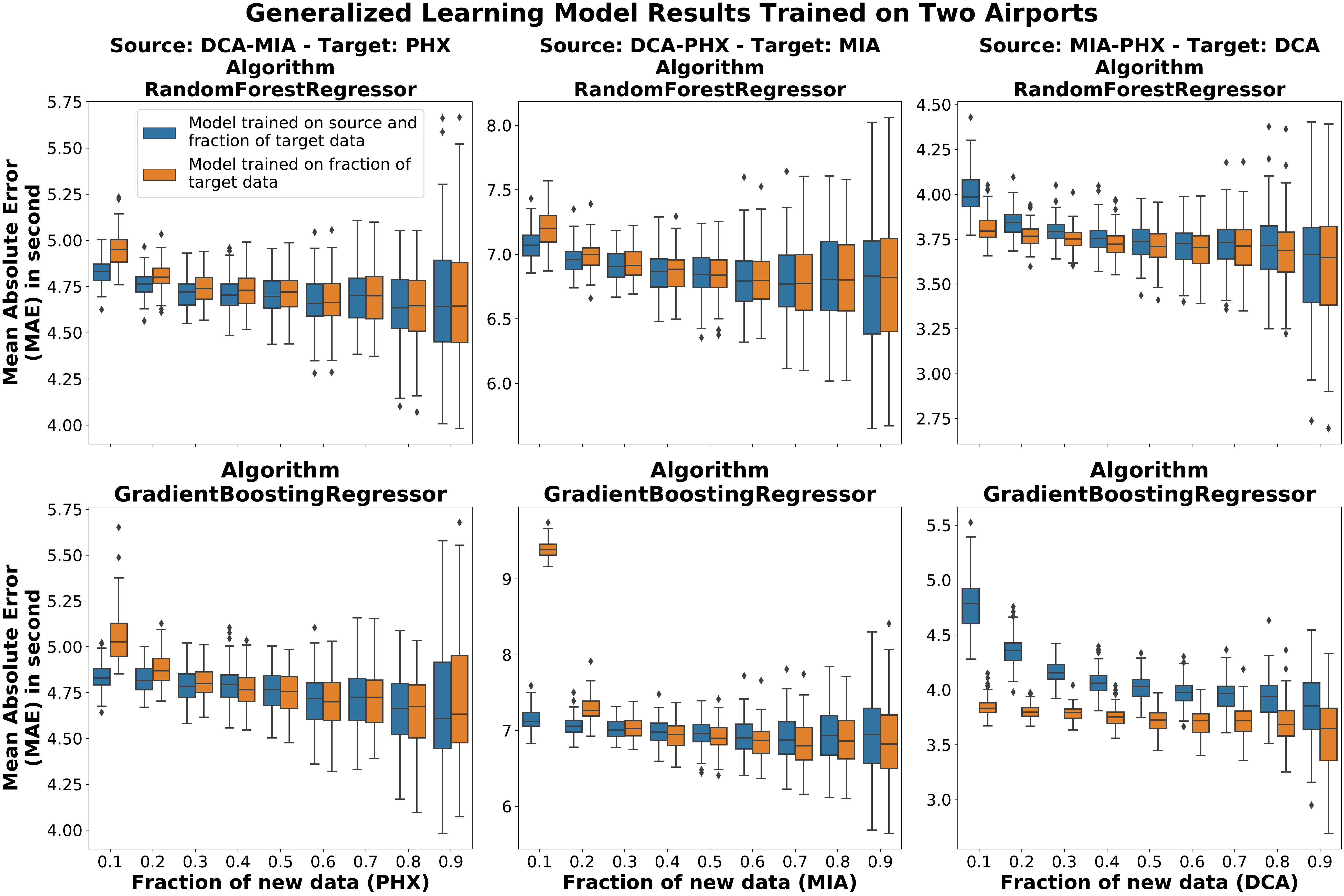}
    \caption{Generalized learning results of two different machine learning algorithms (Random Forest Regressor and Gradient Boosting Regressor), using source data from two different airports. The train data is from two source airports and a fraction ($\alpha = 0.1 \dots 0.9$) of target airport data. Testing is performed on the remainder ($1 - \alpha $) of target airport data. Evaluation is done using Mean Absolute Error in second. Vertical view shows the results of the same set of source - target data with two different machine learning algorithms.}
    \label{fig:generalized-learning-two-sources}
\end{figure*}

In the second version of the generalized learning experiment, we train the Generalized Model on data from two source airports along with a fraction of the data from the target airport, then use that model to make predictions on the remaining target data. Both Random Forest Regressor and Gradient Boosting Regressor show the benefit of generalized learning in 4 out of the 6 considered cases when the fraction of target data is small ($\alpha = 0.1 \dots 0.4$), but the benefit is not evident when the fraction of target data is large ($\alpha = 0.5 \dots 0.9$). However, the benefits, when they are present, are are more obvious than the first version. The results are shown in Fig. \ref{fig:generalized-learning-two-sources}.

The fact that the performance increases with a low percentage of new target airport data gives promising initial results for the model’s capability to track non-stationary airport environments. The generalized learning benefit is not observed for Washington DCA airport, and for two other airports when fraction of target data is large ($\alpha = 0.5 \dots 0.9$), in both versions of the generalized learning experiment, that is, using one or two source airports. This is likely due to the small quantity of data currently available and a combination of the different delay characteristics of those airports (see Fig. \ref{fig:arot-hist}). The feature equivalences technique cannot sufficiently address these differences. A rigorous transfer learning paradigm could be employed. This will be investigated in future work.

\section{Conclusion}
\label{sec:conclusion}
We have shown that by substituting categorical features with numerical equivalences, we can have a prediction model which exhibits generalized learning capability without sacrificing any appreciable performance. The prediction model gives the ATCO awareness of the AROT of upcoming flights 75 to 120 seconds before the aircraft crosses the runway threshold. The prediction model also reduces the uncertainty of the target airport AROT by 32\%-47\%.

We also have shown the benefit of generalized learning in the majority of test cases. This result enables a prediction model to be deployed at a target airport where AROTs data may not be available or difficult to obtain.

A current potential limitation of this research is the small quantity of training data points used, which can lead to skewed assessments in some cases. We envisage addressing this in future work when more data is available.

\section*{Acknowledgment}
This project was partially funded by Saab Singapore Pte. Ltd. The authors would also like to express their gratitude to Robert Brown and David Sargrad at Saab Sensis, Syracuse, NY for providing the Saab Sensis' Aerobahn data.

\bibliographystyle{IEEEtran}
\bibliography{IEEEabrv, ref}

\end{document}